\documentclass[journal]{IEEEtran}
\usepackage{amsthm,amsmath,amssymb}
\usepackage{graphicx}
\usepackage{url}
\usepackage{algorithm}
\usepackage{algorithmic}
\usepackage{color}
\usepackage{verbatim}
\usepackage{float}
\usepackage{pdfpages}
\usepackage{subfigure}
\usepackage{tabularx}
\usepackage{xcolor}
\usepackage[square, numbers, sort]{natbib}
\newtheorem{lemma}{Lemma}
\renewcommand{\algorithmicensure}{ \textbf{Output:}}

\ifCLASSINFOpdf

\else
\fi
\begin{document}
%
\title{Price-Discrimination Game for Distributed Resource Management in Federated Learning}
%
%
%
\author{Han Zhang, Halvin Yang and Guopeng Zhang
\thanks{Han Zhang and Guopeng Zhang are with the School of Computer
Science and Technology, China University of Mining and Technology, Xuzhou
221116, China (e-mail: hanzhangl@cumt.edu.cn; gpzhang@cumt.edu.cn).}
\thanks{Halvin Yang is with the Department of Electronic and Electrical Engineering, University College London, WC1E 7JE London, U.K. (e-mail: uceehhy@ucl.ac.uk).}}

%
%


\maketitle

\begin{abstract}
In federated learning (FL) systems, the parameter server (PS) and clients form a \textit{monopolistic market}, where the number of PS is far less than the number of clients.
To improve the performance of FL and reduce the cost to incentive clients, this paper suggests distinguishing the pricing of FL services provided by different clients, rather than applying the same pricing to them. 
The price is differentiated based on the performance improvements brought to FL by clients and their heterogeneity in computing and communication capabilities.
To this end, a \textit{price-discrimination game} (PDG) is formulated to comprehensively address the \textit{distributed resource management} problems in FL, including \textit{multi-objective trade-off}, \textit{client selection}, and \textit{incentive mechanism}.
As the PDG includes a mixed-integer nonlinear programming problem, a distributed semi-heuristic algorithm with low computational complexity and low communication overhead is designed to solve the Nash equilibrium (NE) of the PDG.
The simulation result verifies that the NE achieves a good tradeoff between the training loss, training time, and the cost of motivating clients to participate in FL.
\end{abstract}


\section{Introduction}
\IEEEPARstart{F}{ederated} learning (FL) allows a parameter server (PS) to communicate with multiple clients to train artificial intelligence (AI) models that adapt to client data without directly accessing the data \cite{2016Communication}.
In addition to privacy protection, FL also needs to address the \textit{distributed resource management} (DRM) issues, including
1) \textit{Multi-objective trade-off}:
how can a PS achieve a good compromise between \textit{performance} (such as the test accuracy) and \textit{efficiency} (such as the training time)?
2) \textit{Client selection}: how can a PS select suitable clients from potential candidates to achieve its goals?
3) \textit{Incentive mechanism}: participating in FL requires the contribution of data, computing and communication resources. How can a PS motivate clients to participate in FL, and how can a selected client allocate its resources to achieve the goal of the PS?

The above issues have been partially addressed in existing work, such as the \textit{client selection} problem studied in \cite{AuctionIncentive} \cite{Adaptive} \cite{Jointly}, the \textit{multi-objective trade-off} problem considered in \cite{Jointly}, and the \textit{incentive mechanisms} designed in \cite{AuctionIncentive} \cite{Motivating} \cite{Task-Load-Aware}.
Specially, ref. \cite{Adaptive} optimizes the allocation of computing resources and amount of training data of clients to improve the testing accuracy. However, this work does not consider the issues of incentive mechanisms.
Ref. \cite{AuctionIncentive} proposes an auction mechanism to incentivize clients to participate in FL. However, this work does not consider other constraints of FL, such as reducing training time.
Ref. \cite{Motivating} uses a Stackelberg game to motivate clients to allocate more computing resources for FL, thereby reducing the training time. However, this work does not consider the issue of client selection.

Furthermore, if the PS and clients are considered as buyers and sellers of FL services respectively, existing methods do not fully utilize the monopolistic market characteristics of FL systems, as the number of PS/buyers is much smaller than the number of clients/sellers.
Most existing works, such as \cite{Crowdsourcing}, provide the same service pricing for different clients. If the PS can differentiate the pricing for services provided by different clients, it can improve the performance of FL and reduce the cost of motivating clients to participate in FL \cite{zhang2011game}.
The price can be differentiated based on the performance improvements brought to FL by clients, and their heterogeneity in computing and communication capabilities.
However, this requires a comprehensive solution to the DRM problems in FL.
Due to the interconnected and tightly coupled nature of these issues, there are significant challenges ahead.

In this paper, we propose a price-discrimination game (PGD) to address the above difficulties.
In the PDG, the PS prices the clients based on the improved \textit{test accuracy} and reduced \textit{training latency} brought by the clients, and the clients are compensated based on their inherent heterogeneity, as clients with different communication and computing capabilities will consume different amounts of resources to achieve the same learning objectives.
By solving the Nash equilibrium (NE) of the PDG, the conflict of interest between the PS and clients can be effectively coordinated, the DRM problem in FL can be comprehensively solved, and the resource utilization of the FL system can be significantly improved.

The main contributions of this paper are as follows: 1) A PDG is proposed to cover all the DRM issues in FL, with the aim of improving the performance of FL and reducing the cost of motivating clients to participate in FL; 2) As the PDG includes a mixed-integer nonlinear programming (MINLP) problem, a distributed iterative algorithm with low computational complexity and low communication overhead is designed to solve the NE of the PDG.
\IEEEpeerreviewmaketitle

\section{System Model}
The considered FL system consists of a PS and a set $\mathcal{M}$ of $|\mathcal{M}|$ clients. The PS is located in an edge server and has direct wireless links with the clients.
Each client $m$ ($\forall m \in \mathcal{M}$) stores a dataset $\mathcal{D}_m$ with $|\mathcal{D}_m|$ data-tuples. A data-tuple $\{x_i, y_i\} \in \mathcal{D}_m$, $\forall i \in \{1,\cdots,|\mathcal{D}_m|\}$, consists of a data sample $x_i$ and its ground-truth label $y_i$.
Let $\omega$ denote the parameter of an AI-model to be trained by the PS. Privacy protection makes the PS unable to directly share the client data. The \texttt{FedAvg} algorithm \cite{2016Communication} can help the PS part solve this problem.

Firstly, the PS selects a subset $\mathcal{N}$ ($\mathcal{N} \subseteq \mathcal{M}$) of $N$ clients from the candidate set $\mathcal{M}$ and broadcasts the global model $\omega$ to the clients. Each selected client $m$ ($\forall m\in\mathcal{N}$) can independently train the localized $\omega$, expressed as $\omega_m$, using dataset $\mathcal{D}_m$.
Let $f(x_i,y_i;\omega_m)$ represent the loss function of the gap between the predicted value $\Tilde{y}_i$ and the truth label $y_i$. The training direction of client $m$ is to minimize the following average prediction loss on the local dataset $\mathcal{D}_m$
\begin{equation}
    F_m(\omega_m)=\frac{1}{|\mathcal{D}_m|}\sum_{\forall \{x_i, y_i\} \in \mathcal{D}_m}f(x_i,y_i;\omega_m),\ \forall m \in \mathcal{N}.
    \label{beforeenergy}
\end{equation}
The training can be implemented by iteratively performing stochastic gradient descent (SGD) on the local dataset for $I_m$ times. After that, client $m$ should upload the trained $\omega_m$ to the PS for aggregation. This ends a round of \textit{local training}.

Upon receiving all the local models uploaded by the clients in set $\mathcal{N}$, the PS can update the global model as 
\begin{equation}
    \omega = \sum_{\forall m \in \mathcal{N}} \frac{|\mathcal{D}_m|}{|\mathcal{D}|} \omega_m,
\end{equation} where $|\mathcal{D}|=\underset{\forall m \in \mathcal{N}}{\sum} |\mathcal{D}_m| $. This ends a round of \textit{global training}.
Such \textit{global training} will be performed iteratively for $I_\text{g}$ rounds upon completion of an FL task.

\subsection{Energy consumption for training $\omega_m$}
Let $c_m$ be the number of CPU cycles required to train $\omega_m$ with one tuple $\{x_i, y_i\}$. Let $f_m^\text{max}$ be the maximum frequency of the CPU of client $m$. 
We use $f_m$, where $0 \leq f_m \leq f_m^\text{max}$, to represent the CPU frequency allocated by client $m$ to train $\omega_m$. Then, in each round of local training, the time spent by client $m$ performing $I_m$ times SGD on $\mathcal{D}_m$ is given by
\begin{equation}
    T^\text{trn}_{m}= \frac{c_m |D_m| I_m}{f_m},\ \forall m \in \mathcal{N}.
\end{equation}
Let $v_m$ denote the effective capacitance coefficient of the CPU of client $m$ \cite{Jointly}. The energy consumed by client $m$ to execute the $I_m$ times SGD is obtained as
\begin{equation}
    E_{m}^\text{trn}=\frac{v_m \left(c_{m} I_m |D_{m}|\right)^3}{(T^\text{trn}_{m})^2 } = v_m f_{m}^2 c_{m} I_m |D_{m}|.
\end{equation} 

\subsection{Energy consumption for uploading $\omega_m$}
Let $p_m$ and $g_m$ be the transmit power of client $m$ and the channel gain from client $m$ to the PS during FL, respectively. The data rate for client $m$ to upload $\omega_m$ to the PS is given by $r_{m}=B\log_2\left(1+\frac{p_{m} g_{m}}{\sigma^2}\right)$, where $B$ is the bandwidth and $\sigma^2$ is the noise power.
The time spent by client $m$ to upload $\omega_m$ to the PS once is given by
\begin{equation}
    T^\text{com}_{m}=  |\omega_m| /r_{m},\; \forall m\in\mathcal{N},
\end{equation}
where $|\omega_m|$ is the size of $\omega_m$.
Therefore, the energy consumed by client $m$ to upload $\omega_m$ to the PS once is given as
\begin{equation}
    E_{m}^\text{com}=p_{m} T^\text{com}_{m}=p_{m}|\omega_m|/{r_{m}},\;\forall m \in \mathcal{N}. 
\end{equation}

\section{Problem Formulation}
Before addressing the DRM issues, the following metrics are provided to measure the performance of FL.

\textit{Test accuracy}:
Let $\omega^*$ denote the model that minimizes the training loss.
Let $\omega_{I_\text{g}}$ denote the model after $I_\text{g}$ rounds of global training.
Then we use $F(\omega^*)$ and $F(\omega_{I_\text{g}})$ to represent the \textit{losses} caused by $\omega^*$ and $\omega_{I_\text{g}}$, respectively.
The \textit{expected difference} between $F(\omega^*)$ and $F(\omega_{I_\text{g}})$ is bounded by \cite{2}
\begin{equation}
     \Gamma =\left(I_\text{g}\underset{{\forall m \in \mathcal{N}}}{\sum} |D_m| \right)^{-\frac{1}{2}}+\left(I_\text{g}\right)^{-1}. \label{accuracy}
\end{equation}
Eq. \eqref{accuracy} indicates that $\Gamma$, i.e., the upper bound of the difference between $F(\omega_{I_g})$ and $F(\omega^*)$ is negatively correlated with the global iteration number $I_g$ of FL and the amount of data used by all clients in the $I_g$ global iterations.

\textit{Training latency}: The global model $\omega$ is updated in a synchronization manner. This means the latency in completing a round of global training, expressed as $T$, depending on the time taken by the last client in set $\mathcal{N}$ to complete a round of local training.
The time used by any client $m$ to complete a round of local training is given by
\begin{equation}
    T_m=T^\text{com}_{m}+T^\text{trn}_{m},\ \forall m \in \mathcal{M}. \label{expression of Tm}
\end{equation}
Then, the expression of $T$ is obtained as
\begin{equation}
    T = \max \left(T_m|\forall m \in \mathcal{N}\right). \label{system completion time}
\end{equation}

To ensure training efficiency, the PS sets a threshold $T_0$ for $T_m$, $\forall m \in \mathcal{M}$.
This means that the following condition
\begin{equation}
    T_m \leq T_0,\ \forall m \in \mathcal{N}.
\end{equation} should be satisfied by any participating client $m$. 
Otherwise, the PS will not make payment to the client.

\subsection{The objective of clients}
The aim of a client is to earn compensation from the PS to offset its resource consumption while making as much extra profit as possible. Therefore, the \textit{client utility} is defined as
\begin{align}
    U_m =\alpha_m (T_0-T_m)-\beta_m(E_{m}^\text{trn}+E_{m}^{\text{com}}),\ \forall m \in \mathcal{M}. \label{utility function of client}      
\end{align} where $\alpha_m$ is the price provided by the PS to client $m$, and $\beta_m$ is the cost per unit of energy consumption of client $m$.

The \textit{first} term of $U_m$ represents the gross profit earned by client $m$ from the PS. The shorter the time $T_m$ it takes for client $m$ to complete a round of local training, the larger $(T_0-T_m)$, and the more revenue the client can earn from the PS.
The \textit{second} term of $U_m$ represents client's resource endowment, i.e., the cost of resources consumed by client $m$ to complete a round of local training using time $T_m$.
The smaller $T_m$, the more energy client $m$ consumes, and the larger $\beta_m$, the more concerned the client is about energy consumption.

The seller's problem for client $m$ is formulated as
\begin{align}
    & \underset{f_m} {\text{max}}\ U_m,\ \forall m \in \mathcal{M}, \label{objective of client} \\
    \mbox{s. t.}\  & U_m=0,\ \forall m \in \mathcal{M}-\mathcal{N}, \tag{\ref{objective of client}.1} \label{constraint of client 1} \\
    & U_m > 0,\ \forall m \in \mathcal{N}, \tag{\ref{objective of client}.2} \label{constraint of client 2} \\
    & 0 \leq f_m \leq f_m^\text{max},\ \forall m \in \mathcal{M}. \tag{\ref{objective of client}.3} \label{maximum frequency of client} 
\end{align}
Constraint \eqref{constraint of client 1} states that if client $m$ is not selected, $U_m=0$, $\forall m \in \mathcal{M}-\mathcal{N}$.
Constraint \eqref{constraint of client 2} states that a selected client cannot be given negative utility. This orders that the PS should set a minimum $\alpha_m$ for client $m$, denoted by $\alpha_m^\text{min}$ ($\alpha_m^\text{min}>0$) to ensure $U_m > 0$ when $f_m>0$, $\forall m \in \mathcal{N}$. The solution of $\alpha_m^\text{min}$ will be discussed in Sec. IV-A.

\subsection{The objective of the PS}
As the buyer of FL services, the PS aims to reduce the expected difference in training loss $\Gamma$ after $I_\text{g}$ rounds of global training, reduce the overall training time $I_\text{g} T$, and at the same time reduce the cost of motivating clients to participate in FL.
The utility of the PS is defined as
\begin{equation}
    Q= \kappa \Gamma + \mu I_\text{g} T+ I_\text{g} \sum_{\forall m \in\mathcal{M}}  \alpha_m (T_0-T_m), \label{aim of PS}    
\end{equation} where $\kappa>0$ and $\mu>0$ are the weighting factors to adjust the importance between the two objectives $\Gamma$ and $I_\text{g}T$.

Let $\mathcal{A}=\{\alpha_1,. . . ,\alpha_|\mathcal{_M}_|\}$.
We define that $ \alpha_m=0,\ \forall m \in \mathcal{M}-\mathcal{N}$, and $\alpha_m \geq \alpha_m^\text{min},\;\forall m \in \mathcal{N}$. The buyer's problem for the PS is formulated as 
\begin{align}
    & \underset{\mathcal{A},\ \mathcal{N}}{\text{min}}\ Q \label{objective of PS} \\
    \mbox{s. t.}\ 
    &\mathcal{N} \subseteq \mathcal{M} \label{client number limit} \tag{\ref{objective of PS}.1}, \\
    & \alpha_m=0,\ \forall m \in \mathcal{M}-\mathcal{N} \label{price limit 1} \tag{\ref{objective of PS}.2}, \\
    & \alpha_m \geq \alpha_m^\text{min},\;\forall m \in \mathcal{N}. \label{price limit 2} \tag{\ref{objective of PS}.3}
\end{align}

\section{Solving The Game}
\subsection{Optimal strategies for the clients}
For any client $m$ who does not participate in FL, i.e., $\forall m \in \mathcal{M}-\mathcal{N}$, we have $f_m=0$, and according to constraint \eqref{price limit 1}, we also know $\alpha_m=0$. By substituting $f_m=0$ and $\alpha_m=0$ into eq. \eqref{utility function of client}, we have $U_m=0$, $\forall m \in \mathcal{M}-\mathcal{N}$. This means that constraint \eqref{constraint of client 1} is always satisfied.  

For each selected client $m$, $\forall m \in \mathcal{N}$, 
it is easy to prove that $U_m$ is concave w.r.t $f_m$, $\forall f_m > 0$. Therefore, $U_m$ satisfies the property of diminishing marginal utility, and there exists a unique solution to problem \eqref{objective of client}, $\forall f_m > 0$, which is given by
\begin{equation}
    \tilde{f}_m=\left({\frac{\alpha_m}{2\beta_m v_m}}\right)^{1/3},\ \forall m \in \mathcal{N}.   \label{eq:optimal fm for client without power limit} 
\end{equation}
Due to constraint \eqref{maximum frequency of client}, the optimal solution to problem \eqref{objective of client} is finally obtained as
\begin{equation}
    f_m^*=\text{min}\left(\tilde{f}_m,\ f_m^\text{max}\right),\ \forall m \in \mathcal{N}. \label{eq:optimal fm for UE}
\end{equation}

However, $f_m^*$ cannot guarantee $U_m > 0$, $\forall m \in \mathcal{N}$. In order to meet constraint \eqref{constraint of client 2}, the condition $\alpha_m > \alpha_m^\text{min}$ ($\forall m \in \mathcal{N}$) should be satisfied, where $\alpha_m^\text{min}>0$ is lowest price that the PS can offer to client $m$. Due to the convexity of $U_m$, we can solve the function $ U_m =0$ and obtain
\begin{equation}
    \alpha_m^\text{min}=\frac{\beta_m\left(E_m^\text{trn}+E_m^\text{com} \right)} {T_0-T_m},\ \forall m \in \mathcal{N}. \label{minimum alpha}
\end{equation} 


\subsection{Optimal strategy for the PS}
Problem \eqref{objective of PS} is an MINLP problem and is difficult to solve due to the client selection constraint \eqref{client number limit} and the uncertain training latency $T$ given in eq. \eqref{system completion time}. Because it cannot be solved by conventional methods, we propose a semi-heuristic approach to obtain a high-quality solution to the PDG. Before presenting the algorithm, the following \textit{Lemma} is given.

\begin{lemma}
    Regardless of the maximum limit of $f_m$ in constraint \eqref{maximum frequency of client}, the optimal time for client $m$ ($\forall m \in \mathcal{N}$), to complete a session of local training satisfies
\begin{equation}
     T_m=T = \max \left(T_m|\forall m \in \mathcal{N}\right),\;\forall m \in \mathcal{N}.
\end{equation}
\end{lemma}
\begin{proof}
    If $T_m<T$ ($\exists m \in \mathcal{N}$), the PS can decrease $Q$ by reducing $\alpha_m$ to make $T_m=T$, without affecting the strategies and utilities of other clients. \label{lemma1}
\end{proof}
The proposed algorithm for solving the PDG includes the following two steps:

\textbf{Step 1:} \textit{Solve problems \eqref{objective of client} and \eqref{objective of PS} without considering constraint \eqref{client number limit}, assuming that $\mathcal{N}$ is known.}

By substituting eq. \eqref{eq:optimal fm for client without power limit} into eq. \eqref{expression of Tm}, we can get
\begin{equation}
    \tilde{\alpha}_m=\frac{2c_m^3 I_m^3 D_m^3 r_m^3 \beta_m v_m}{(T_m r_m-|\omega|)^3},\ \forall m \in \mathcal{N},\label{eq:alpha1}
\end{equation}
which is the price offered by the PS to client $m$ when client $m$ uses strategy $\tilde{f}_m$. To ensure a positive utility for client $m$, the PS must guarantee $\tilde{\alpha}_m\geq\alpha_m^\text{min}$. By solving the function $\tilde{\alpha}_m=\alpha_m^\text{min}$, we can obtain the solution of $T_m$ denoted by $\tilde{T}_m$.

Now, we consider constraint \eqref{maximum frequency of client}. Because $T_m$ decreases monotonically w.r.t $f_m$, we can substitute $f_m=f_m^\text{max}$ into eq. \eqref{expression of Tm} and get the solution of $T_m$, denoted by $T_m^\text{min}$. Note that $T_m^\text{min}$ is the shortest time required for client $m$ to complete a round of local training.

If $f_m^\text{max}<\tilde{f}_m$, then $T_m^\text{min}>\tilde{T}_m$.
It means that client $m$ cannot obtain a positive $U_m$, even with $f_m=f_m^\text{max}$. To ensure $U_m>0$, the PS needs to increase $\alpha_m$ in the \textit{first} term of eq. \eqref{utility function of client}. Then, the value space of $T_m$ is given by
\begin{equation}
    T_m^\text{min} \leq T_m \leq \text{max}\left(\tilde{T}_m,T_m^\text{min}\right),\ \forall m \in \mathcal{N}. \label{valute space of Tm}
\end{equation}

Following \textbf{Lemma 1} and eq. \eqref{valute space of Tm}, we know that
\begin{equation}
    T_m=\left\{
    \begin{aligned}
        &T,\; T_m^\text{min} \leq T \leq \tilde{T}_m,\\
        &\text{max}\left(\tilde{T}_m,T_m^\text{min}\right),\; T > \tilde{T}_m,
    \end{aligned}
    \right.\ \forall m \in \mathcal{N},\label{new expression of Tm}
\end{equation}
and the corresponding price provided by the PS to client $m$ is
\begin{equation}
    \alpha_m=\left\{
    \begin{aligned}
        &\tilde \alpha_m ,\; T_m^\text{min} \leq T \leq \tilde{T}_m,\\
        &\alpha_m^\text{min},\; T > \tilde{T}_m.
    \end{aligned}
    \right. \ \forall m \in \mathcal{N}. \label{alpha_m}
\end{equation}

Based on eqs. \eqref{new expression of Tm} and \eqref{alpha_m}, we can rewrite the utility of the PS in eq. \eqref{aim of PS} as
\begin{equation}
    \hat{Q}(T)=\kappa \Gamma + \mu I_\text{g} T+ I_\text{g} \sum_{\forall m \in\mathcal{N}} R_m(T), \label{eq:Q rewrite}
\end{equation} where
\begin{equation}
    R_m(T)=\left\{
    \begin{aligned}
        &\tilde\alpha_m\left(T_0-T\right),\; T_m^\text{min} \leq T \leq \tilde{T}_m,\\
        &\alpha_m^\text{min}\left(T_0-\text{max}\left(\tilde{T}_m,T_m^\text{min}\right)\right),\; T > \tilde{T}_m,
    \end{aligned}
    \right.\label{payment}
\end{equation} is the payment that client $m$ can receive from the PS.
$\hat{Q}(T)$ is continuous and has a finite number of non-derivable points w.r.t $T$. Furthermore, $\hat{Q}(T)$ also has the following property:
\begin{lemma}
    When $\alpha_m>\alpha_m^{\min}$, $\forall m \in \mathcal{N}$, $\hat{Q}(T)$ is monotonic or decreases first and then increases w.r.t $T$.
\end{lemma}
\begin{proof}
When $\alpha_m>\alpha_m^\text{min}$, the value space of $T$ is given by $\text{max}(T_m^\text{min}|\forall m \in \mathcal{N}) \leq T \leq \text{max} (\tilde T_m|\forall m \in \mathcal{N})$.
The first derivative of $\hat{Q}$ w.r.t $T$ is obtained as
\begin{equation}
    \frac{\partial \hat{Q}}{\partial T}=I_{\text{g}}\mu+I_{\text{g}}\sum_{m \in \mathcal{N}} \frac{\partial \hat{R}_m}{\partial T} \label{eq:16}
\end{equation}where 
\begin{equation}
\small
    \frac{\partial \hat{R}_m}{\partial T}=\left\{
    \begin{aligned}
        &-\beta v_m \left(\frac{c_m I_m D_m}{T-T_m^\text{com}}\right)^3\left(1+ \frac{3(T_0-T)}{T-T_m^\text{com}}\right),\; T_m^\text{min} \leq T \leq \tilde{T}_m,\\
        &0,\; T > \tilde{T}_m.
    \end{aligned}
    \right. \label{deritive of Rm}
\end{equation}

\normalsize
From eq. \eqref{deritive of Rm}, one can derive that $\frac{\partial^2 \hat{R}_m}{\partial T^2} \geq 0$ and $\frac{\partial^2 \hat{Q}}{\partial T^2} \geq 0$ for $T_m^\text{min} \leq T \leq \tilde{T}_m$. This means that $\frac{\partial \hat{Q}}{\partial T}$ is monotonically increasing w.r.t $T$ for $T_m^\text{min} \leq T \leq \tilde{T}_m$. 
Form eq. \eqref{deritive of Rm}, it is also known that $\frac{\partial \hat{R}_m}{\partial T}  \to -\infty$ as $T \to T_m^\text{com}$.
Because $\frac{\partial^2 \hat{R}_m}{\partial T^2} \geq 0$, $\frac{\partial \hat{R}_m}{\partial T}$ is monotonically increasing w.r.t $T$ as $T \to T_m^\text{com}$.
From the above analysis, we know that $\frac{\partial \hat{Q}}{\partial T}$ is always negative or first negative and then positive w.r.t $T$. Therefore, $\hat{Q}$ is monotonic or decreases first and then increase w.r.t $T$ when $\alpha_m >\alpha_m^{\min}$. 
\end{proof}

From \textbf{Lemma 2}, we know that $\hat{Q}(T)$ has a unique minimum in the feasible domain, and it can be obtained by continuously increasing $\alpha_m$ of client $m$ with the largest $T_m$ to motivating it to reduce $T_m$, until $T$ and $\hat{Q}(T)$ cannot be further reduced. Then, the following algorithm is designed to solve problems \eqref{objective of client} and \eqref{objective of PS} with a known $\mathcal{N}$. 

\begin{algorithm}
\caption{Solve problems \eqref{objective of client} and \eqref{objective of PS} with a known $\mathcal{N}$.}
\label{alg:Framwork1}
\begin{algorithmic}[1]
\renewcommand{\algorithmicrequire}{\textbf{Input:}}
\renewcommand{\algorithmicensure}{\textbf{Output:}}
\FOR{each client $m$, $\forall m \in \mathcal{N}$}
    \STATE Obtain $T_m^\text{min}$ by using eq. \eqref{expression of Tm}, obtain $\tilde{T}_m$ and $\alpha_m$ corresponding to $\tilde{T}_m$ by using eqs. \eqref{minimum alpha} and \eqref{eq:alpha1}, respectively, and upload $\alpha_m^\text{min}$ and $\tilde{T}_m$ to the PS.
\ENDFOR
\STATE 
The PS initializes $\alpha_m$ and $T_m=\tilde{T}_m$, and obtains $\tilde Q$ by using eq. \eqref{eq:Q rewrite}. 
\REPEAT
\STATE The PS increases the payment $\alpha_m$ to client $m$ with the largest $T_m$ and feedback the updated $\alpha_m$ to client $m$. 
\STATE With the updated $\alpha_m$, each client $m$ updates $f_m$ and $T_m$ by using eqs. (\ref{eq:optimal fm for UE}) and (\ref{expression of Tm}), respectively, and uploads $T_m$ to the PS.
\STATE The PS update $\hat{Q}$ by using eq. \eqref{eq:Q rewrite}.
\UNTIL $\hat{Q}$ start to increase or $T=\max\left(T_m^\text{min}|\forall m \in \mathcal{N}\right)$. 
\end{algorithmic}
\end{algorithm}
Since the result $\mathcal{A}=\{(\alpha_1, f_1) \cdots (\alpha_{|\mathcal{N}|}, f_{|\mathcal{N}|})\}$ obtained by executing \textbf{Algorithm 1} solves both problems \eqref{objective of client} and \eqref{objective of PS}, it constitutes the Nash equilibrium (NE) of the game.
In the NE state, no one can benefit from unilateral strategic changes when the strategies of others remain unchanged.

\textbf{Step 2:} \textit{Apply client selection constraint \eqref{client number limit} to the solution obtained in \textit{Step 1}}. We adopt the \textit{Greedy Algorithm} (GA) to solve this problem. The main idea is to allow the PS to select all the candidates in set $\mathcal{M}$ to participate in FL at the beginning. Then, the PS constantly removes client $m$ that has the maximum marginal loss and causes high cost $Q$ of the PS, until $Q$ cannot be further reduced by eliminating any client or the number of selected clients reaches a certain threshold $|\mathcal{N}|=N_0$. The overall algorithm to solve the proposed PDG is concluded in the following \textbf{Algorithm 2}.

\begin{algorithm}
\caption{Overall algorithm to solve the proposed PDG.}
\label{alg:Framwork2}
\begin{algorithmic}[1]
\renewcommand{\algorithmicrequire}{\textbf{Input:}}
\REQUIRE $\mathcal{M}$, the candidate client set for the PS. \\
\renewcommand{\algorithmicensure}{\textbf{Output:}}
\ENSURE The selected client set $\mathcal{N}$, the optimal strategy $f_m^*$ of client $m$, $\forall m \in \mathcal{N}$, and the optimal strategy $\mathcal{A^*}=\{\alpha_1^*,. . . ,\alpha_{|\mathcal{_M}|}^*\}$ of the PS. \\
\STATE \textbf{Initialize} $\mathcal{N}=\mathcal{M}$.
\STATE Obtain $Q'=\hat{Q}$ by performing \textbf{Algorithm 1} on set $\mathcal{N}$.
\REPEAT
\FOR{ each client $m$, $\forall m \in \mathcal{N}$}
\STATE Let $\mathcal{N}_{-m}=\mathcal{N} \backslash m$.
\STATE Obtain $Q_{-m}=\hat{Q}$ by performing \textbf{Algorithm 1} on set $\mathcal{N}_{-m}$.
\ENDFOR \\
\STATE Let $m'= \text{argmin}_{\forall m \in \mathcal{N}}\left(Q_{-m}|\forall m \in \mathcal{N}\right)$.
\STATE Update $\mathcal{N}=\mathcal{N}_{-m'}$, and $Q'=Q_{-m'}$.
\UNTIL{$Q'$ starts to increase or the number of the selected clients satisfies $|\mathcal{N}| \leq N_0$.}
\end{algorithmic}
\end{algorithm}

The computational complexity of \textbf{Algorithm 2} is analyzed as follows. In line 6 of \textbf{Algorithm 2}, \textbf{Algorithm 1} is executed repeatedly to update the utility $Q$ of the PS. The convergence of \textbf{Algorithm 1} mainly depends on strategy interaction between the PS and clients in lines 5-9 of \textbf{Algorithm 1}. Considering that the clients can use convex optimizer to obtain $T_m^\text{min}$ and $\tilde{T}_m$, the time complexity of \textbf{Algorithm 1} is $\mathcal{O}(l |\mathcal{M}|)$, where $\mathcal{O}(l)$ is the complexity of the convex optimizer. Therefore, the time complexity of \textbf{Algorithm 2} is $\mathcal{O}(l|\mathcal{M}|^2)$.As the algorithms are executed before the start of the training, it is fully compatible with existing FL framework. The main challenge of the algorithms comes from the price update steps 3)-10) of \textbf{Algorithm 2}, as they need to be iteratively executed to find the optimum of utilities of the PS and clients.
Since the algorithms are performed by the PS and the strategies of the clients are solved in closed form, they satisfy the constraint of FL on the low computing power of clients. In addition, the information that needs to be transferred between the PS and the clients is the service price $\alpha_m$ and the training latency $T_m$, which brings small communication overhead to the system.

\section{Simulation Results}
In the simulation, the PS trained the convolutional neural network (CNN) \texttt{AlexNet} using the CIFAR-10 dataset, which consists of 60000 32x32 color images, divided into 10 classes.
The size of \texttt{AlexNet} is $|\omega|=0.6$ Mb.
In the independent identical distributed (IID) setting, the data-tuples of each class are evenly distributed across $|\mathcal{M}|$ = 40 clients, and $|\mathcal{D}_m|\sim \mathcal{U}(100, 2000)$, where $\mathcal{U}(\cdot)$ represents uniform distribution. In the Non-IID setting,  the number of classes selected by each client follows a uniform distribution $\mathcal{U}(1, 10)$, and the number of data-tuples in each selected class also follows a uniform distribution $\mathcal{U}(10, 400)$.  
The weighting factors in eq. \eqref{aim of PS} are set to $\kappa=10^6$ and $\mu=1$, and the latency threshold is set to $T_0=10$ s.
For the clients, the switch capacitance coefficient is set to $v_m = 10^{-28}$ \cite{Jointly}, the number of CPU cycles required to train one data-tuple $c_m = 5 \times 10^5$, the maximum CPU frequency $f_m^\text{max} \sim \mathcal{U}(2,4)$ GHz, the cost per unit of energy consumption $\beta_m=1$, the bandwidth $B=1$ MHz, and the maximum transmit power $p_m\sim \mathcal{U}(0.02,0.1)$ W.
The channel power gain is set to $g_m = 0. 001/{d_m^3}$, where $d_m \sim \mathcal{U}(10, 100)$ (in meters) is the distance between client $m$ and the PS, and the noise power at the PS is $\sigma^{2} = 10^{-9}$.

To verify the performance of the proposed PDG, the following algorithms have also been implemented.
Given the threshold $|\mathcal{N}|=N_0$ for the number of clients that the PS can select from set $\mathcal{M}$, the vanilla \texttt{FedAvg} randomly selects $N_0$ clients from $\mathcal{M}$ to participate in FL.
The optimized \texttt{FedAvg} with client selection \cite{Hybrid-FL} can choose $N_0$ clients with the maximum number of data-tuples to achieve the highest test accuracy (referred to as the Accuracy First Client Selection Algorithm (ACA)) or choose $N_0$ clients with the shortest $T_m$ to minimize the overall training time (referred to as the Time First Client Selection Algorithm (TCA)).

\begin{figure}[!htb]
    \centering
    \includegraphics[width=0.4\textwidth]{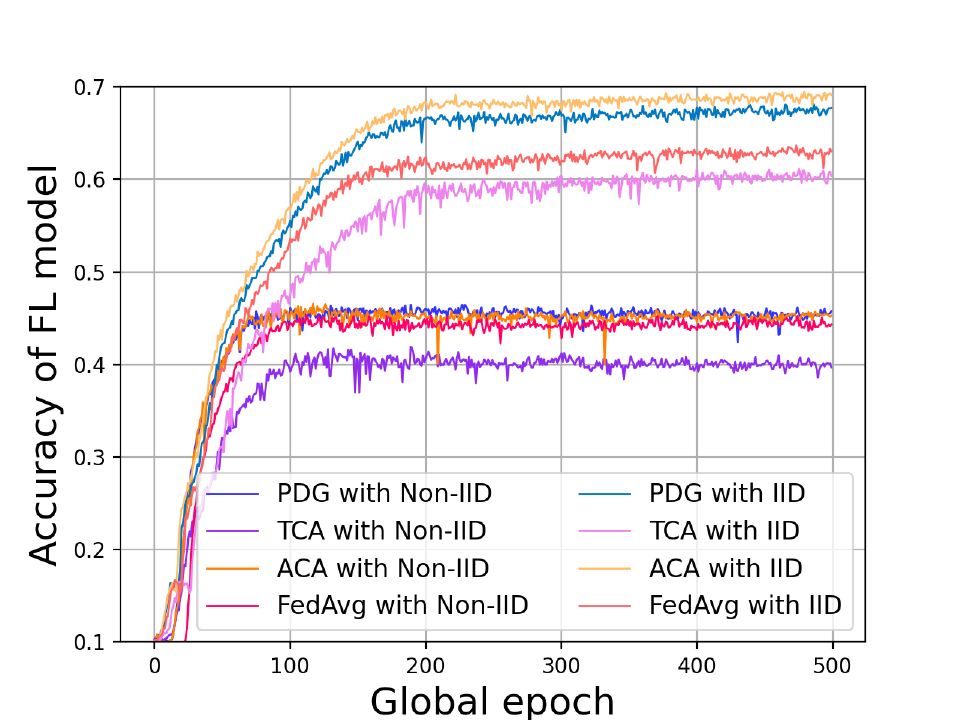}
    \caption{The convergence of accuracy.}
    \label{acc}
\end{figure}

Setting $N_0=10$, Fig. 1 shows the convergence of \textit{test accuracy} for \texttt{AlexNet} trained using different algorithms under IID and Non-IID settings, respectively. From Fig. \ref{acc}, we see that as the number of global training increases, the \textit{test accuracy} achieved by the PDG is very close to that achieved by the AFA, and is significantly superior to the other algorithms under both the IID and Non-IID settings.



\begin{figure}[!htb]
    \centering
    \subfigure[The training time.]{\includegraphics[width=0.23\textwidth]{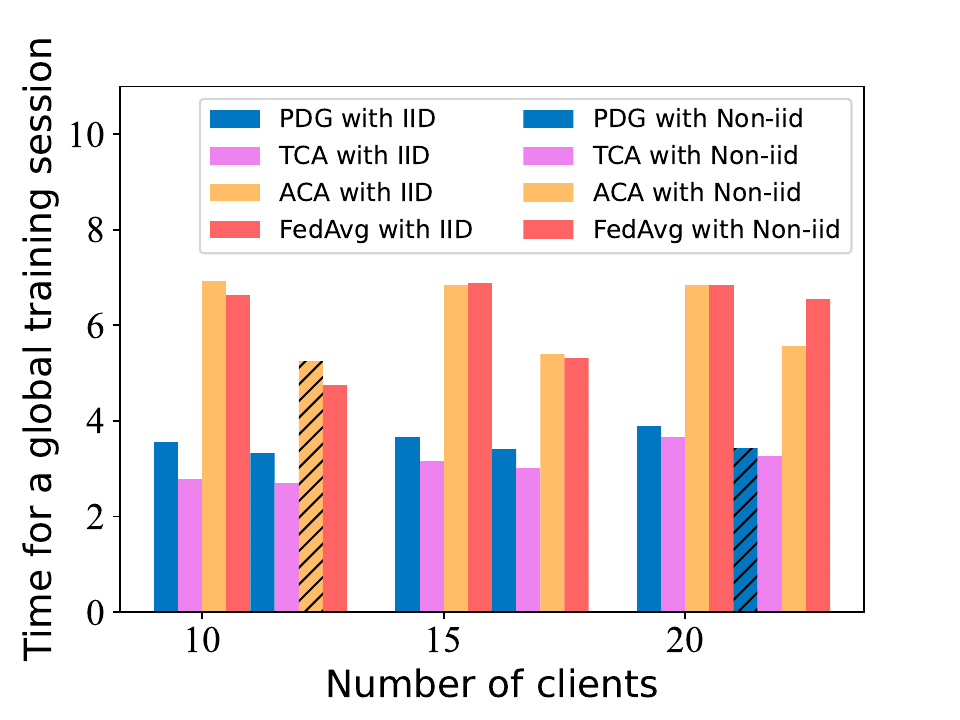}\label{time}}
    \subfigure[The utility of the PS.]{\includegraphics[width=0.24\textwidth]{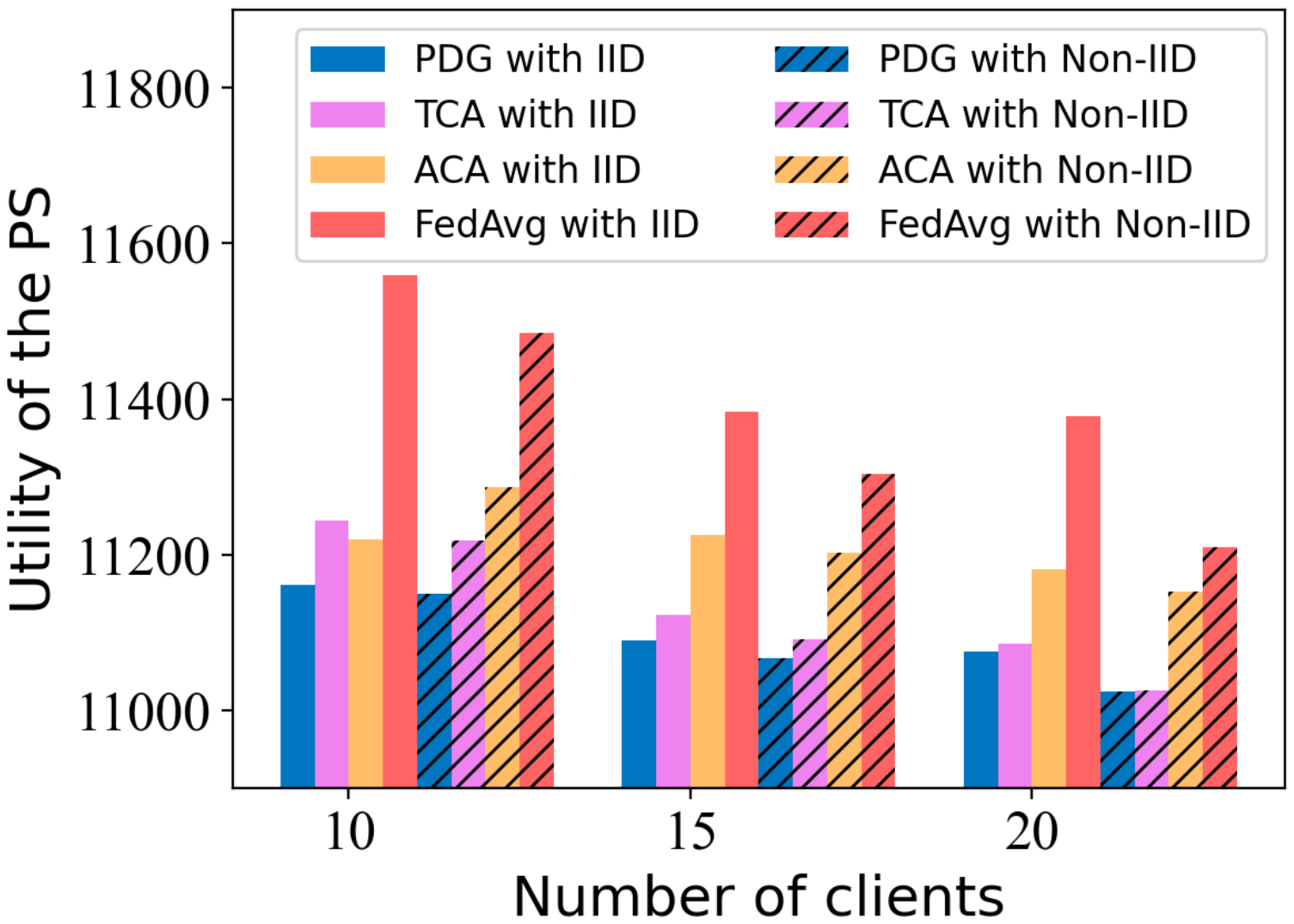}\label{loss}}
    \caption{Performance of client selection algorithm.}
    \label{fig: figure}
\end{figure}

Next, we set $N_0$ to 10, 15 and 20, respectively. Figs. \ref{time} and \ref{loss} compare the \textit{training time} and \textit{utility} obtained by the PS using different algorithms under IID and Non-IID settings, respectively. From Fig. \ref{time}, we see that the PDG consumes slightly more time than the TCA under both IID and Non-IID settings, but is much lower than the other algorithms. Due to the ACA achieving the highest \textit{test accuracy}, while the TCA minimizing the overall training time, it is verified that the proposed PDG can achieve a good tradeoff between \textit{performance} and \textit{efficiency}. This conclusion was also validated in the results shown in Fig. \ref{loss}, as in all cases, the PDG brings the PS the lowest utility. Note that the utility of the PS is a multi-objective composite function composed of training loss, training time, and the cost of motivating clients to participate in FL. The smaller the $Q$, the better the performance.
 
When the PS selects 10 clients from the potential client set to participate in FL, the following Fig. \ref{utility1} provides a comparison of the utility achieved by each client when employing different algorithms.
\begin{figure}[!htb]
    \centering
    \includegraphics[width=0.4\textwidth]{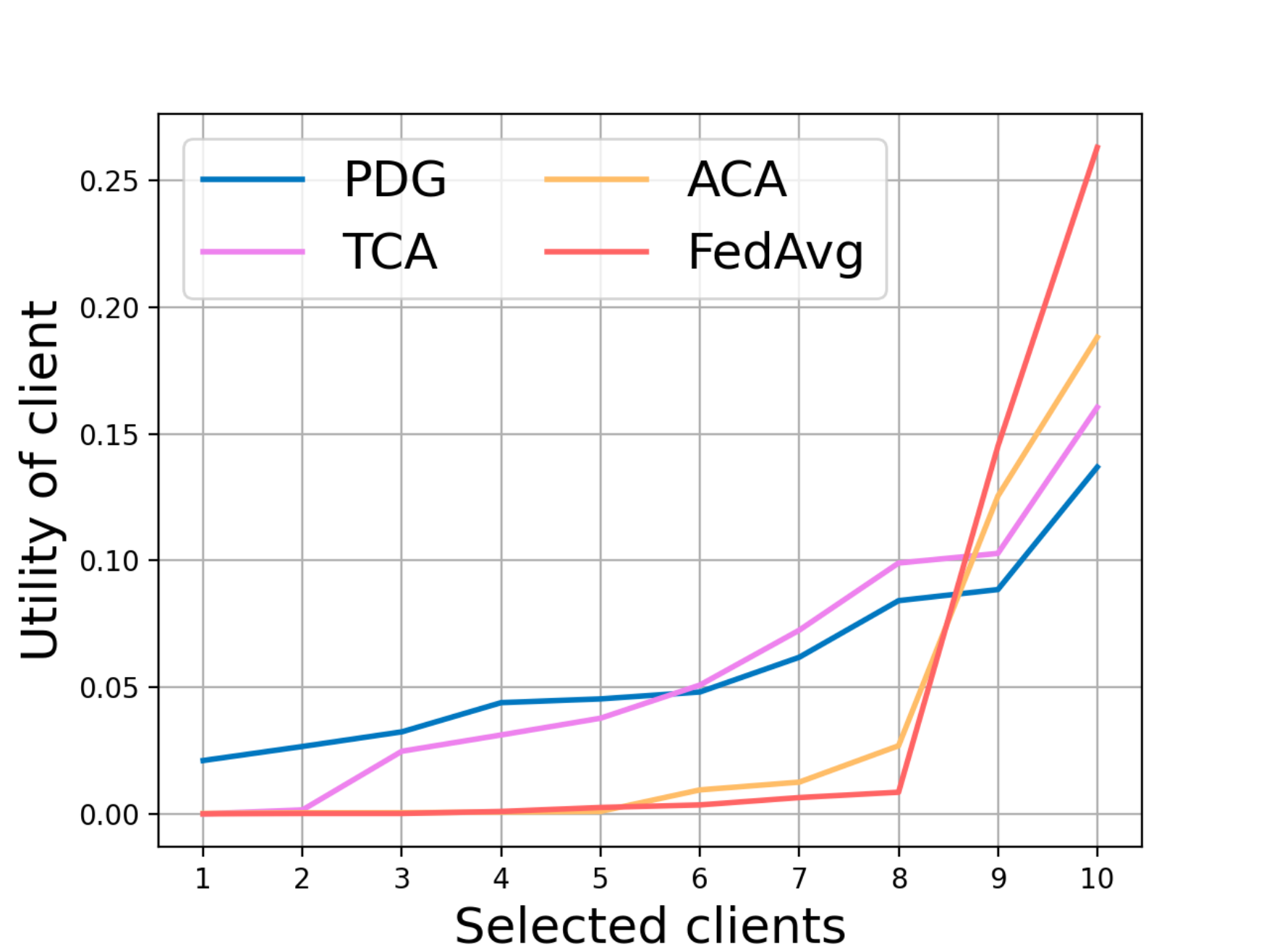}
    \caption{The utility achieved by clients using different algorithms.}
    \label{utility1}
\end{figure}

From Fig. \ref{utility1}, it can be observed that the proposed PDG minimizes the differences in utility obtained by each client. This is because the PDG has the capability to select clients that are most suitable for the requirements of FL tasks, namely, the selected clients exhibit minimal differences in computing capabilities and communication conditions. Consequently, the variations in utility among each selected client are also minimized. In contrast, clients chosen by other algorithms exhibit instances of utility close to zero, and some clients achieve significantly higher utility compared to others. This is attributed to the larger differences in computing resources, communication conditions, or training workloads among clients selected by these alternative algorithms.

\begin{figure}[!htb]
    \centering
    \includegraphics[width=0.4\textwidth]{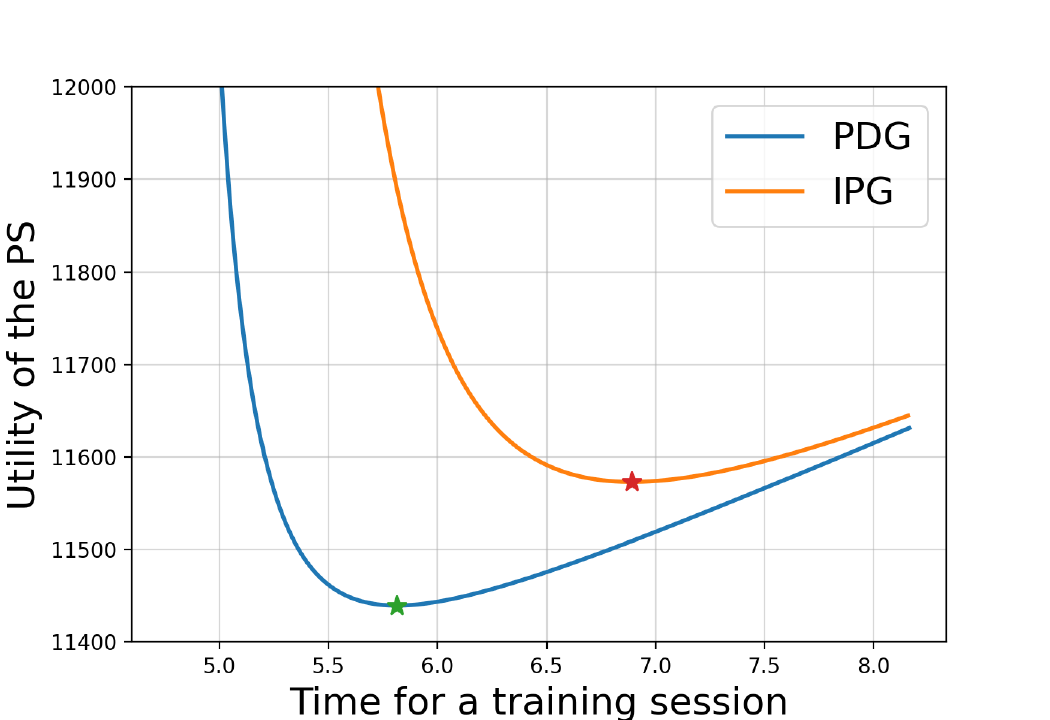}
    \caption{The utility of the PS with different pricing strategy.}
    \label{pricing}
\end{figure}

Finally, we validate the benefits of the proposed \textit{price discrimination strategy} on FL.
To this end, we implemented the algorithm based on the \textit{identical pricing strategy} \cite{Crowdsourcing}, which is referred to as the \textit{identical pricing game} (IPG). The IPG only prices the services provided by different clients based on the performance improvements they bring to FL, without considering the heterogeneity of their resource endowments. 
Under the IID settings, we designate 10 clients to participate in FL and continuously increase the time $T$ (see eq. \eqref{system completion time}) required for the clients to complete a round of global training. As $T$ increases, Fig. \ref{pricing} compares the utilities achieved by the PS in the PDG and IPG.

From Fig. \ref{pricing}, we can see that for any $T$, the PDG is superior to the IPG because it always achieves lower utility than the IPG. As shown by the asterisk in Fig. \ref{pricing}, the minimum utility (i.e. the best performance of the PS) that the PDG can achieve is better than what the IPG can achieve. This is because the PS knows the resource endowment of the client through the PDG, thereby gaining a bargaining advantage in the monopolistic market and achieving lower utility than using the IPG.

\begin{figure}[!htb]
    \centering
    \includegraphics[width=0.4\textwidth]{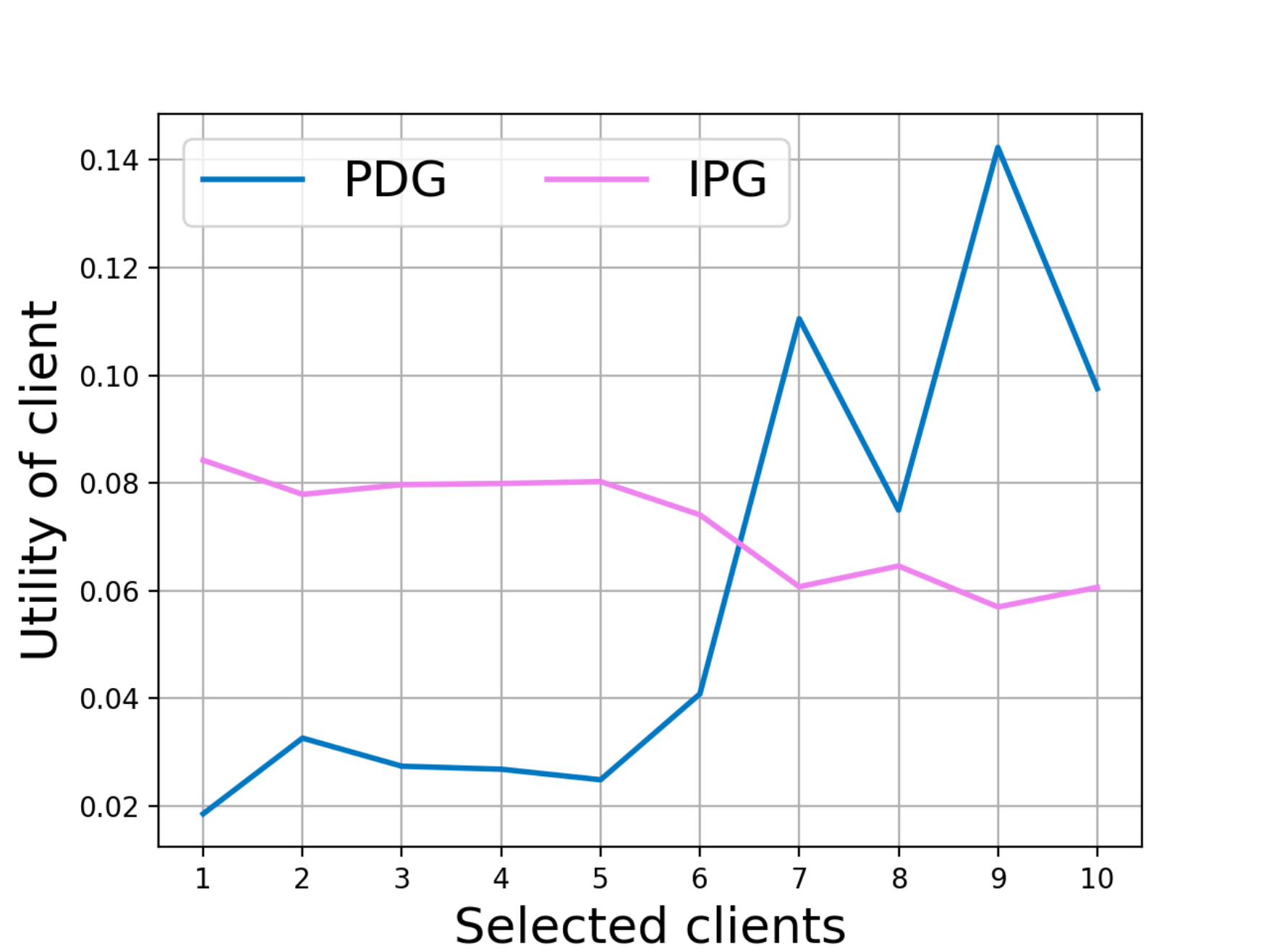}
    \caption{The utility of clients with different pricing strategy.}
    \label{utility2}
\end{figure}

Next, we compared the utility obtained by each client under different pricing strategies in the game. The simulation results are presented in Fig. \ref{utility2}.

In Fig. \ref{utility2}, the training workloads (i.e., training data amounts) of clients 1-10 are sorted from the smallest to the largest. It can be observed that when using the PDG method, the utility of clients 7-10, who train more data, is significantly higher than that of clients 1-6. This is because the PDG can distinguish the contributions of different clients in FL, meaning that when using the same local training time, clients training more data can obtain higher rewards. The IPG is not sensitive to the contributions of clients in FL, and a simple uniform pricing strategy may give an advantage to clients with lower training workloads, which is unfair to clients that train more data.

\section{Conclusions and Future Directions}
This paper explores a single PS coordinating multiple clients for model training. Our subsequent research will further investigate the problem of non independent and identically distributed (NIID) client data, and extend the monopoly model to a oligopoly model with multiple PSs as resource buyers. The aim is to enhance the bargaining power of clients as resource sellers in the FL market.

\textit{It should be noted that implementing the incentive mechanism for federated learning using the proposed price discrimination game will face the following problems. From eq. (12.2), we know that once the price offered by the PS to client $m$ results in $U_m > 0$, even if only slightly larger than zero, e.g. $U_m \to 0^+$ in the analytical sense, client $m$ is motivated to contribute its resources to the FL task. This results in clients being unable to guarantee appropriate returns for their contribution. Even a slightly positive price will incentivize the seller to participate in the FL, regardless of how much it benefits the buyer. The main reasons for the above results are as follows.}

\begin{itemize}
    \item[1)] \textit{The above results are attributed to the inherent characteristics of the price game model. Price gaming originates from the analysis of the demand and prices of buyers and sellers in microeconomic markets. Particularly, in the monopolistic market model adopted in this paper, where there is only one buyer, it creates the possibility for buyers to make purchases without incurring any cost. In a buyer monopolistic market, the quantity purchased is determined by the marginal value and marginal cost of the buyer monopolist. In this article, the buyer's marginal value (i.e., the reduced utility per unit time) is determined by the buyer's (PS) demand for training time $\mu$, while the marginal cost (i.e., the cost required to shorten unit time) increases as training time decreases. When the marginal utility of shortening training time (i.e., the difference between marginal value and marginal cost) equals zero, it represents the maximum purchase quantity for the buyer.}
    
    \item[2)] \textit{Another direct reason for the above results is the assumption of a monopolistic market. This makes the seller's zero profit equilibrium unaffected by market size. As the number of clients increases, each client can only receive an infinitely small price (in analytical situations) or the minimum positive price allowed by the quantitative system (in practical situations), while the profit obtained by the buyer PS is increasing linearly. In extreme cases, where the number of sellers increases from $n$ to infinite, we will see that each seller does not have or receives very little profit, but the buyer receives unlimited profit.}
\end{itemize}

\textit{At present, as far as the author knows, there are several ways to solve the above problems.}

\begin{itemize}
    \item[1)] \textit{Replace the monopoly model with a oligopoly model. In an oligopoly market, there may be only a few buyers in the market. Regardless of the number of sellers, as long as there is more than one buyer, there exists price competition among the buyers in the market. Competition among buyers will lead to higher prices, which is advantageous for sellers.}

    \item[2)] \textit{This problem is also due to the utility function of the parameter server (Equation (7)) employed in the paper not taking into account the diversity of user data. When the client's data exhibits a non-iid distribution, the data from each client can be viewed as non-substitutable goods, or the substitutability is reduced. This, in turn, weakens competition among sellers and enhances the utility for the seller. Therefore, the scenario of non-iid data is worth exploring in depth. It is worth noting that even without replacing the oligopoly model with a monopoly model, the above problems can be partially alleviated when considering the irreplaceability of client data. }
\end{itemize}


\ifCLASSOPTIONcaptionsoff
  \newpage
\fi
\footnotesize
\bibliographystyle{IEEEtran}
\bibliography{cite}

\end{document}